\documentclass{article}

\PassOptionsToPackage{numbers, compress}{natbib}

\usepackage[final]{nips_2018}

\usepackage[utf8]{inputenc} 
\usepackage[T1]{fontenc}    
\usepackage{hyperref}       
\usepackage{url}            
\usepackage{booktabs}       
\usepackage{amsfonts}       
\usepackage{amsmath}
\usepackage{nicefrac}       
\usepackage{microtype}      
\usepackage{xcolor}
\usepackage{graphicx}
\usepackage{multirow}

\title{Bayesian Sparsification of \\ Gated Recurrent Neural Networks}

\author{
  Ekaterina Lobacheva$\bf{}^{1}$\thanks{Equal  contribution}\;,\quad Nadezhda Chirkova$\bf{}^{1}$\footnotemark[1]\;,\quad Dmitry Vetrov$\bf{}^{1,2}$\\
  ${}^1$Samsung-HSE Laboratory, National Research University Higher School of Economics\\
	${}^2$Samsung AI Center \hspace{35pt} Moscow, Russia\\
  \texttt{\{elobacheva, nchirkova, dvetrov\}@hse.ru} \\
}

\begin{document}

\maketitle

\begin{abstract}
  Bayesian methods have been successfully applied to sparsify weights of neural networks and to remove structure units from the networks, e. g. neurons. 
  We apply and further develop this approach for gated recurrent architectures. Specifically, in addition to sparsification of individual weights and neurons, we propose to sparsify preactivations of gates and information flow in LSTM. It makes some gates and information flow components constant, speeds up forward pass and improves compression. Moreover, the resulting structure of gate sparsity is interpretable and depends on the task.
\end{abstract}

\section{Introduction}
Recurrent neural networks (RNNs) yield high-quality results in many applications~\cite{las,hypernet,qa,trans} but often overfit due to overparametrization. In many practical problems, RNNs can be compressed orders of times with only slight quality drop or even with quality improvement~\cite{emnlp, pruning, groupsparseLSTM}. Methods for RNN compression can be divided into three groups: based on matrix factorization~\cite{kroneker, tjandra}, quantization~\cite{binary} or sparsification~\cite{emnlp, pruning, groupsparseLSTM}.

We focus on RNNs sparsification. Two main groups of approaches for sparsification are pruning and Bayesian sparsification. In pruning~\cite{pruning, groupsparseLSTM}, weights with absolute values less than a predefined threshold are set to zero. Such methods imply a lot of hyperparameters (thresholds, pruning schedule etc). Bayesian sparsification techniques~\cite{dmolch,lognormal,chris,l0,emnlp} treat weights of an RNN as random variables and approximate posterior distribution over them given sparsity-inducing prior distribution. 
After training weights with low signal-to-noise ratio are set to zero. 
This allows eliminating the majority of weights from the model without time-consuming hyperparameters tuning. Also, Bayesian sparsification techniques can be easily extended to permanently set to zero intermediate variables in the network’s computational graph~\cite{lognormal, chris} (e.g. neurons in fully-connected networks or filters in convolutional networks). It is achieved by multiplying such a variable on a learnable weight, finding posterior over it and setting the weight to zero if the corresponding signal-to-noise ratio is small.

In this work, we investigate the last mentioned property for gated architectures, particularly for LSTM. Following~\cite{emnlp, dmolch}, we sparsify individual weights of the RNN. Following~\cite{chris}, we eliminate neurons from the RNN by introducing multiplicative variables on activations of neurons. Our main contribution is the introduction of multiplicative variables on preactivations of the gates and information flow in LSTM. This leads to several positive effects. Firstly, when some component of preactivations is permanently set to zero, the corresponding gate becomes constant. It simplifies LSTM structure and speeds up computations. Secondly, we obtain a three-level hierarchy of sparsification: sparsification of individual weights helps to sparsify gates and information flow 
(make their components constant), and sparsification of gates and information flow helps to sparsify neurons 
(remove them from the model). As a result, the overall compression of the model is higher.

\section{Preliminaries}
Consider a dataset of $N$ objects $(x_i, y_i)$ and 
a model $p(y|x, W)$ parametrized by a neural network with weights $W$.
In~\cite{dmolch}, the authors propose a Bayesian technique called Sparse variational dropout (SparseVD) for neural networks sparsification. This model comprises log-uniform prior over weights: $p(|w_{ij}|) \propto \frac 1 {|w_{ij}|}$ and fully factorized normal approximate posterior: $q(w_{ij}) = \mathcal{N}(w_{ij}| m_{ij}, \sigma^2_{ij})$.
To find parameters of the approximate posterior distribution, evidence lower bound (ELBO) is optimized:
\begin{equation}
\label{elbo}
\sum_{i=1}^N \mathbb{E}_{q(W)} \log p(y^i|x^i, W) - KL(q(W)||p(W)) \rightarrow \max_{m, \sigma}
\end{equation}
Because of the log-uniform prior, for the majority of weights signal-to-noise ratio 
$m^2_{ij}/\sigma^2_{ij} \rightarrow 0$ and these weights do not affect network's output. 
In~\cite{emnlp} SparseVD is adapted to RNNs.

In~\cite{chris} the authors propose to multiply activations of neurons on group variables $z$ and to learn and sparsify group variables along with $W$.
They put standard normal prior on $W$ and log-uniform prior on $z$. The first prior moves mean values of $W$ to 0, and it helps to set to zero $z$ and to remove neurons from the model. This model is equivalent to multiplying rows of weight matrices on group variables.

\section{Proposed method}
To sparsify individual weights, we apply SparseVD~\cite{dmolch} to all weights of the RNN, taking into account recurrent specifics underlined in~\cite{emnlp}. To compress layers and remove neurons, we follow~\cite{chris} and introduce group variables for the neurons of all layers (excluding output predictions), and specifically, $z^x$ and $z^h$ for input and hidden neurons of LSTM. 

The key component of our model is introducing groups variables $z^i,\, z^f, \, z^g, \, z^o$ on preactivations of gates and information flow. The resulting LSTM layer looks as follows:
\begin{align}
& f = ~ \sigma \biggl( \bigl(W^h_f (h_{t-1} \odot z^h) + W^x_f (x_t \odot z^x) \bigr)  \odot z^f  +b_f \biggr)
\quad \text{\{same for $i$, $o$ and $g$\}} \\
& c_t = f \odot c_{t-1} +  i \odot  g \quad 
~~~h_t =  o \odot tanh(c_t) 
\end{align}
Described model is equivalent to multiplying rows and columns of weight matrices on group variables: 
\[
\hat w^h_{f, ij} = w^h_{f, ij} ~\cdot z^h_{i} ~\cdot z^f_{j}
\quad \text{\{same for $i$, $o$ and $g$\}}
\]
We learn group variables $z$ in the same way as weights $W$: approximate posterior with fully factorized normal distribution given fully factorized log-uniform prior distribution\footnote{Our experiments show that log-uniform prior on individual weights gives better results than standard normal prior used in~\cite{chris}.}. To find approximate posterior distribution, we maximize ELBO~\eqref{elbo}. After learning, we set all weights and group variables with signal-to-noise ratio less than 0.05 to 0.

If some component of $z^i, z^f, z^o, z^g$ is set to 0, the corresponding gate or information flow component becomes constant (equal to activation function of bias). It means that we don't need to compute this component, and the forward pass through LSTM is accelerated.

\paragraph{Related work.} 
In~\cite{groupsparseLSTM} the authors propose a pruning-based method that removes neurons from LSTM and argue that independent removing of $i, \,f, \,g,\, o$ components may lead to invalid LSTM units. In our model, we do not remove these components but make them constant,  gaining compression and acceleration with correct LSTM structure.

\section{Experiments}
\begin{table*}[ht!]
  \normalsize
  \centering
  \begin{tabular}{clcccc}
Task & Method & Quality & Compression  & Neurons $x$ - $h$ & Gates\\ 
\hline
&\,Original  & 84.1 & 1x & $300 - 128$ & 512 \\
IMDb&\,SparseVD W  & {\bf 84.47} & 1135x & $7 - 9$ & 22\\ 
\multicolumn{1}{c}{\textcolor{gray}{Accuracy \%}\!\!}
&\,SparseVD W+N  & 83.98 & 17874x & $1-5$ & $12$\\ 
&\,SparseVD W+G+N  & 83.98 & {\bf19747x} & $\bf 1 - 4$ & {\bf 6}\\ 
\hline
&\,Original  & {\bf 90.6} & 1x & $300 - 512$& $2048$\\
AGNews&\,SparseVD W  & 89.01 & 350x &$193 - 65$& $ 72$ \\
\multicolumn{1}{c}{\textcolor{gray}{Accuracy \%}\!\!}
&\,SparseVD W+N  & 88.55 & 645x & $43-17$ &  $62$ \\ 
&\,SparseVD W+G+N  & 88.41 & {\bf647x}  & $\bf 43 - 14$  & {\bf39}\\ 
\hline
&\,Original  & $1.499 - 1.454$ & 1x & $50 - 1000$ &4000\\
Char PTB&\,SparseVD W  & $1.472 - 1.429$ &  7.9x  & $50 - 431$& 1718\\
\multicolumn{1}{c}{\textcolor{gray}{Bits-per-char}\!\!}
&\,SparseVD W+N  & $1.478 -  1.430$ & \bf10.2x & $\bf 50-390$& {\bf 1560}\\ 
\multicolumn{1}{c}{\textcolor{gray}{Valid-Test}\!\!}
&\,SparseVD W+G+N  & $\bf 1.467 -  1.425$ & 9.8x & $50-404$& 1563\\

\hline
&\,Original  & $135.6 -129.3$ & 1x & $10000- 256$& 1024\\
Word PTB&\,SparseVD W & $\bf 115.0-109.2$& 22.1x & $9990 -156$& 288\\
\multicolumn{1}{c}{\textcolor{gray}{Perplexity}\!\!}
&\,SparseVD W+N  & $116.1-111.0$ & 23.5x & $9992-127$& 276\\ 
\multicolumn{1}{c}{\textcolor{gray}{Valid-Test}\!\!}
&\,SparseVD W+G+N  & $122.0-116.5$ & {\bf 24.5x} & $\bf 9972-117$& {\bf 201}\\ 
\end{tabular}
\caption{Quantitative results. Compression is equal to $|w|/|w\neq0|$. In last two columns number of remaining neurons and non-constant gates in the recurrent layer are reported.}\label{tab:class}
\end{table*}
\vspace{-3mm}
\begin{figure}[h]
    \centering
        \begin{tabular}{ccc}
        \includegraphics[height=2cm]{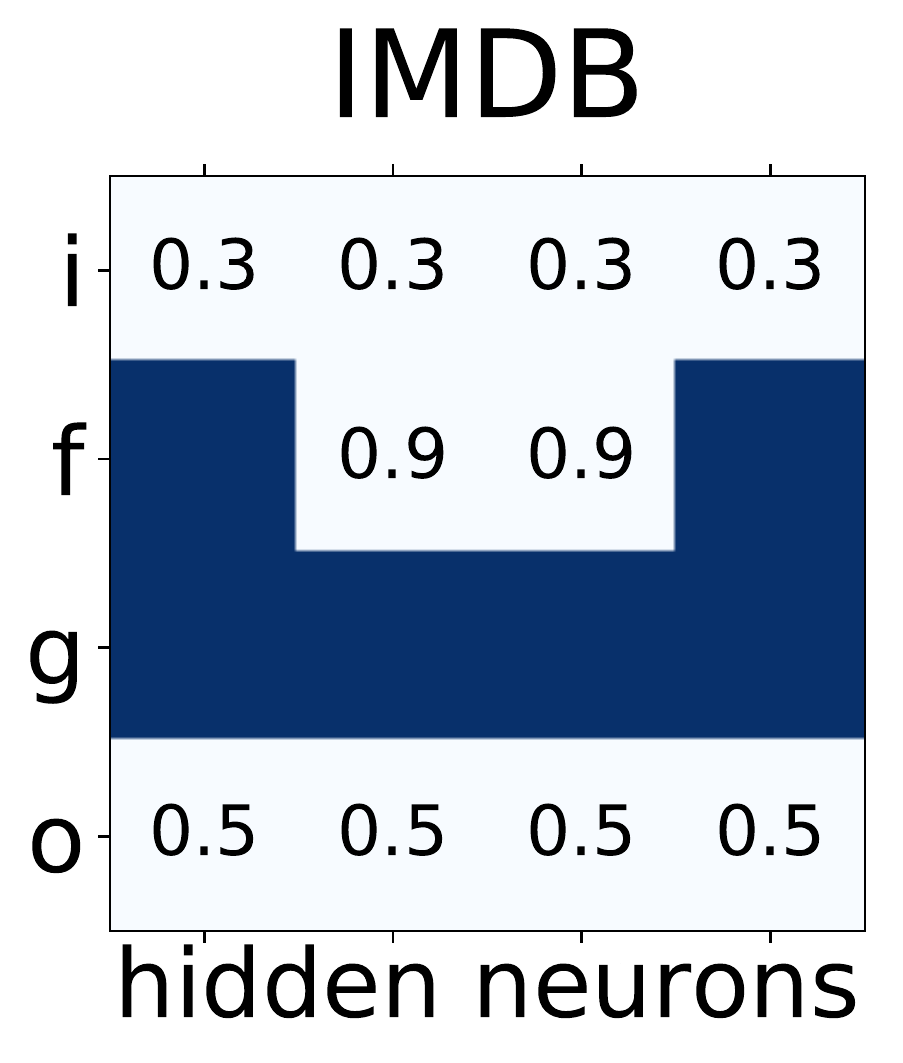}&
           \includegraphics[height=2cm]{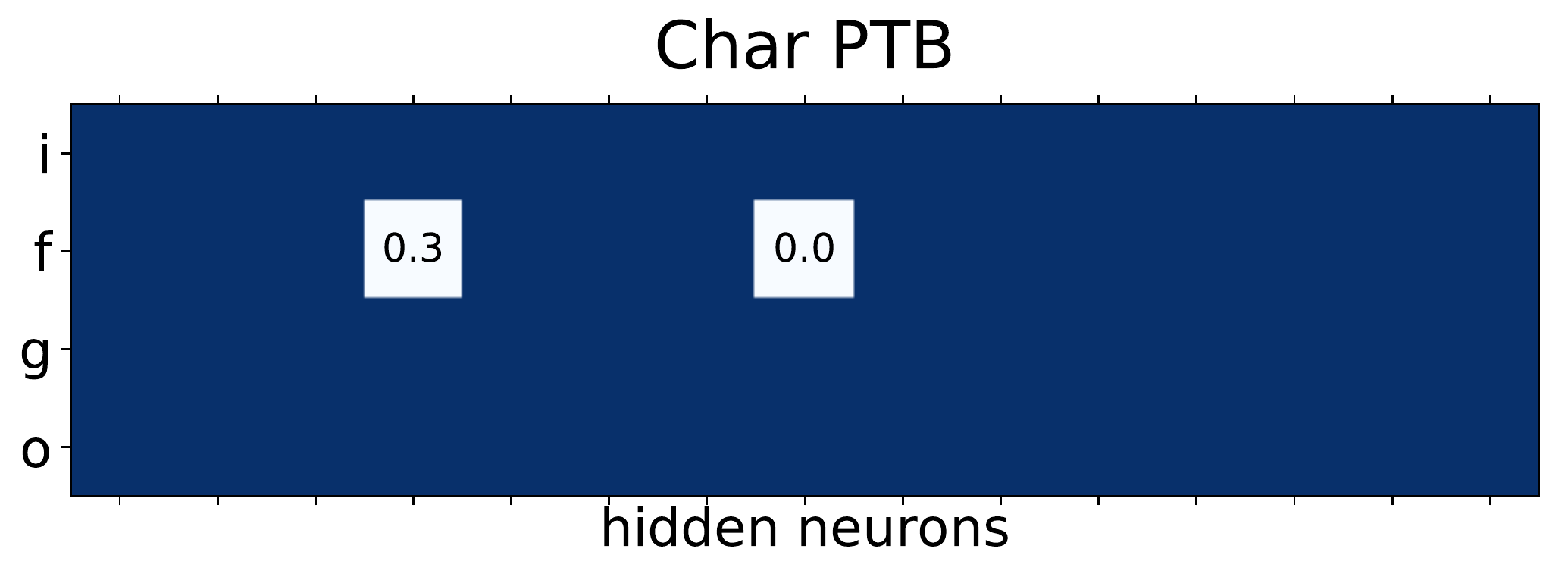} 
           & \includegraphics[height=2cm]{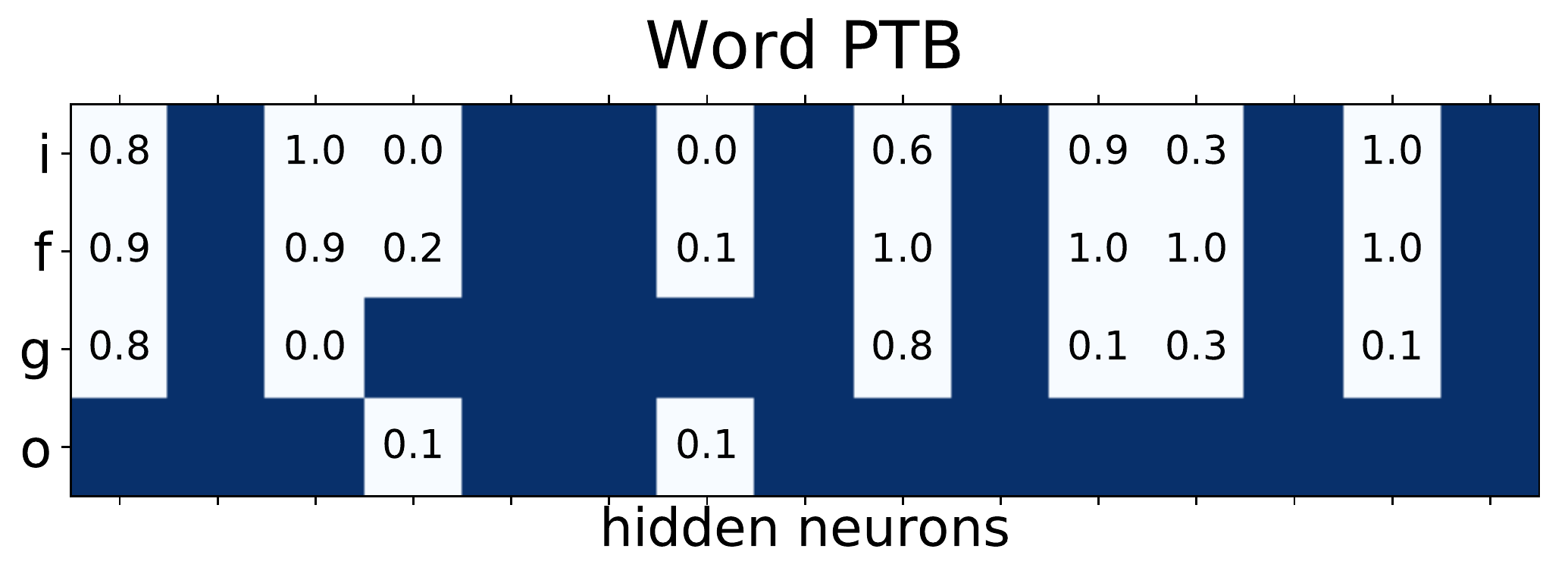} \\
        \end{tabular}
        \caption{Structure of gate sparsity. Non-constant gates are dark blue, constant ones are light blue. Numbers correspond to activation values of constant gates. For language modeling only 15 randomly chosen active neurons are presented.}
        \label{fig:words}
        \vspace{-3mm}
\end{figure}

We perform experiments with LSTM architecture on two types of problems: text classification (datasets IMDb~\cite{IMDB} and AGNews~\cite{agnews}) and language modeling (dataset PTB~\cite{ptb}, character and word level tasks). For text classification, we use networks with an embedding layer, one recurrent layer and an output dense layer at the last timestep. For language modeling, we use networks with one recurrent layer and an output dense layer. More details about network architectures and experiment setup are presented in Appendix~\ref{setup}. 

We compare four models in terms of quality and sparsity: baseline model without any regularization, standard SparseVD model for weights sparsification only (W), SparseVD model with group variables for neurons sparsification (W+N) and SparseVD model with group variables for gates and neurons sparsification (W+G+N). In all SparseVD models, we sparsify weights matrices of all layers. Since in text classification tasks usually only a small number of input words are important, we use additional multiplicative weights to sparsify the input vocabulary in case of group sparsification (W+N, W+G+N) following~\cite{emnlp}. 
On the contrary, in language modeling tasks all input characters or words are usually important, therefore we do not use $z^x$ for this task. Additional sparsification of input neurons in this case noticeably damage models quality and sparsity level of hidden neurons.
To measure the sparsity level of our models we calculate the compression rate of individual weights as follows: $|w|/|w\neq0|$. 
To compute the number of remaining neurons or non-constant gates we use corresponding rows/columns of $W$ and corresponding weights $z$ if applicable. 

Quantitative results are shown in Table~\ref{tab:class}. Multiplicative variables for neurons boost group sparsity level without a significant quality drop. Additional variables for gates and information flow not only make some gates constant but also increase group sparsity level even further. 
Moreover, for a lot of constant gates bias values tend to be very large or small making corresponding gates either always open or close.

Proposed gate sparsification technique also reveals an interesting work-flow structure of LSTM networks for different tasks. Figure~\ref{fig:words} shows typical examples of gates of remaining hidden neurons. For language modeling tasks output gates are very important because models need both store all the information about the input in the memory and output only the current prediction at each timestep. On the contrary, for text classification tasks models need to output the answer only once at the end of the sequence, hence they do not really use output gates. Also, the character level language modeling task is more challenging than the word level one: the model uses the whole gating mechanism to solve it. We think this is the main reason why gate sparsification does not help here.

\section*{Acknowledgments}
The study has been supported by Russian Science Foundation (grant 17-71-20072) and Samsung Research, Samsung Electronics.

\bibliographystyle{apalike}
\bibliography{example_paper}

\appendix
\section{Experimental setup} \label{setup}

{\bf Datasets.} To evaluate our approach on text classification task we use two standard datasets: IMDb dataset~\cite{IMDB} for binary classification and AGNews dataset~\cite{agnews} for four-class classification. We set aside 15\% and 5\% of training data for validation purposes respectively. For both datasets, we use the vocabulary of 20,000 most frequent words. To evaluate our approach on language modeling task we use the Penn Treebank corpus~\cite{ptb} with the train/valid/test partition from~\cite{mikolov11}. The dataset has a vocabulary of 50 characters or 10,000 words.

{\bf Architectures for text classification.} We use networks with one embedding layer of 300 units, one LSTM layer of 128 / 512 hidden units for IMDb / AGNews, and finally, a fully connected layer applied to the last output of the LSTM. Embedding layer is initialized with word2vec~\cite{NIPS2013_5021} / GloVe~\cite{pennington2014glove} and SparseVD models are trained for 800 / 150 epochs on IMDb / AGNews. Hidden-to-hidden weight matrices $W^h$ are initialized orthogonally and all other matrices are initialized uniformly using the method from~\cite{pmlr-v9-glorot10a}. We train our networks using Adam~\cite{adam} with batches of size 128 and a learning rate of 0.0005. Baseline networks overfit for all our tasks, therefore, we  present results for them with early stopping.

{\bf Architectures for language modeling.} To solve character / word-level tasks we use networks with one LSTM layer of 1000 / 256 hidden units and fully-connected layer with softmax activation to predict next character or word. We train SparseVD models for 250 / 150 epochs on character-level / word-level tasks. All weight matrices of the networks are initialized orthogonally and all biases are initialized with zeros. Initial values of hidden and cell elements are not trainable and equal to zero. For the character-level task, we train our networks on non-overlapping sequences of 100 characters in mini-batches of 64 using a learning rate of 0.002 and clip gradients with threshold 1. For the word-level task, networks are unrolled for 35 steps. We use the final hidden states of the current mini-batch as the initial hidden state of the subsequent mini-batch (successive mini batches sequentially traverse the training set). The size of each mini-batch is 32. We train models using a learning rate of 0.002 and clip gradients with threshold 10. Baseline networks overfit for all our tasks, therefore, we  present results for them with early stopping.

{\bf Sparsification.} For all weights that we sparsify, we initialize $\log\sigma$ with -3. We eliminate weights with signal-to-noise  ratio less then $\tau=0.05$.

\end{document}